\documentclass[10pt,twocolumn,letterpaper]{article}

\usepackage{cvpr}              % To produce the CAMERA-READY version
\usepackage{multirow}
\usepackage{wrapfig}
\usepackage{graphicx}
\usepackage{amsmath}
\usepackage{amssymb}
\usepackage{booktabs}
\usepackage{multirow}
\usepackage{wrapfig}
\usepackage{makecell}
\definecolor{myred}{HTML}{9C1629}
\definecolor{mygreen}{HTML}{50812D}
\definecolor{myblue}{HTML}{DAE3F5}
\usepackage[table]{xcolor}
\usepackage{array}
\definecolor{cvprblue}{rgb}{0.21,0.49,0.74}
\usepackage[pagebackref,breaklinks,colorlinks,allcolors=cvprblue]{hyperref}

\def\paperID{6797} % *** Enter the Paper ID here
\def\confName{CVPR}
\def\confYear{2026}

\title{Decision Boundary-aware Generation for Long-tailed Learning}

\author{{Jiacheng Yang$^1$}\quad{Ruichi Zhang$^1$}\quad{Chikai Shang$^1$}\quad{Mengke Li$^2$}\\{Xinyi Shang$^3$}\quad{Junlong Gao$^1$}\quad{Yonggang Zhang$^4$}\quad{Yang Lu$^{1}$\thanks{Corresponding Author: Yang Lu (luyang@xmu.edu.cn).}}\\
\footnotesize $^1$Key Laboratory of Multimedia Trusted Perception and Efficient Computing, Ministry of Education of China, Xiamen University, Xiamen, China\\\footnotesize$^2$College of Computer Science and Software Engineering, Shenzhen University, Shenzhen, China\\\footnotesize$^3$Department of Statistical Science, University College London, London, UK\\\footnotesize$^4$Division of Arts and Machine Creativity, Hong Kong University of Science and Technology, Hong Kong, China\\
\tt\small 23020250157840@stu.xmu.edu.cn\quad zhangruichi@stu.xmu.edu.cn\quad 
ckshangl2@gmail.com\\
\tt\small 
mengkeli@szu.edu.cn\quad 
xinyi.shang.23@ucl.ac.uk\quad
jlgao@xmu.edu.cn\quad 
zhangyg@ust.hk\\
\tt\small 
luyang@xmu.edu.cn}

\begin{document}
\maketitle
% \footnotetext[1]{Key Laboratory of Multimedia Trusted Perception and Efficient Computing, Ministry of Education of China, Xiamen University, Xiamen, China.}
% \footnotetext[2]{Shenzhen University, Shenzhen, China.}
% \footnotetext[3]{University College London, London, UK.}
% \footnotetext[4]{Division of Arts and Machine Creativity, Hong Kong University of Science and Technology, Hong Kong, China.}
\renewcommand{\thefootnote}{}
\footnotetext{Accepted by CVPR 2026}
\begin{abstract}
Long-tailed data bias decision boundaries toward head classes and degrade tail class accuracy. Diffusion-based generative augmentation address this problem by generating additional data, while head-to-tail transfer further mitigate the generator bias inherit from long-tailed dataset. However, we show that while head-to-tail transfer helps balance the decision space of the classifier, it also induces latent non-local feature mixing that entangles inter-class features, causing decision boundary overlap and tail class distribution shift. To address this, we first identify the problem of boundary ambiguity and then propose Decision Boundary-aware Generation (DBG) framework, which promotes near-boundary representation learning by generating informative near-boundary samples. Overall, DBG rebalances the long-tailed dataset while yielding more separable decision space for long-tailed learning. Across standard long-tailed benchmarks, DBG consistently improves tail class and overall accuracy with less inter-class overlap. The code of DBG is available at \color{blue}{https://github.com/keepdigitalabc-svg/DBG}.
\end{abstract}
\section{Introduction}
\label{sec:intro}

Vision-based deep learning methods have delivered substantial gains in convenience and efficiency~\cite{fei2021jointly, melzi2024frcsyn, joseph2021towards, deng2023prompt, xu2022hisa, luo2025long}, yet their performance critically depends on the quality and quantity of the training data~\cite{hestness2017deep, kandpal2023large}. In practice, datasets often exhibit long-tailed class distributions, characterized by a few head classes dominating the samples while many tail classes are underrepresented. Models trained on such imbalanced data tend to learn biased decision boundaries that favor head classes and degrade tail class accuracy~\cite{zhang2023deep}. Though many methods have been proposed for long-tailed learning~\cite{ho2024long, ao2025comparative}, the core challenge remains the scarcity of tail samples which shifts decision boundaries. 

\begin{figure}
\centering
{\includegraphics[width=0.48\textwidth]{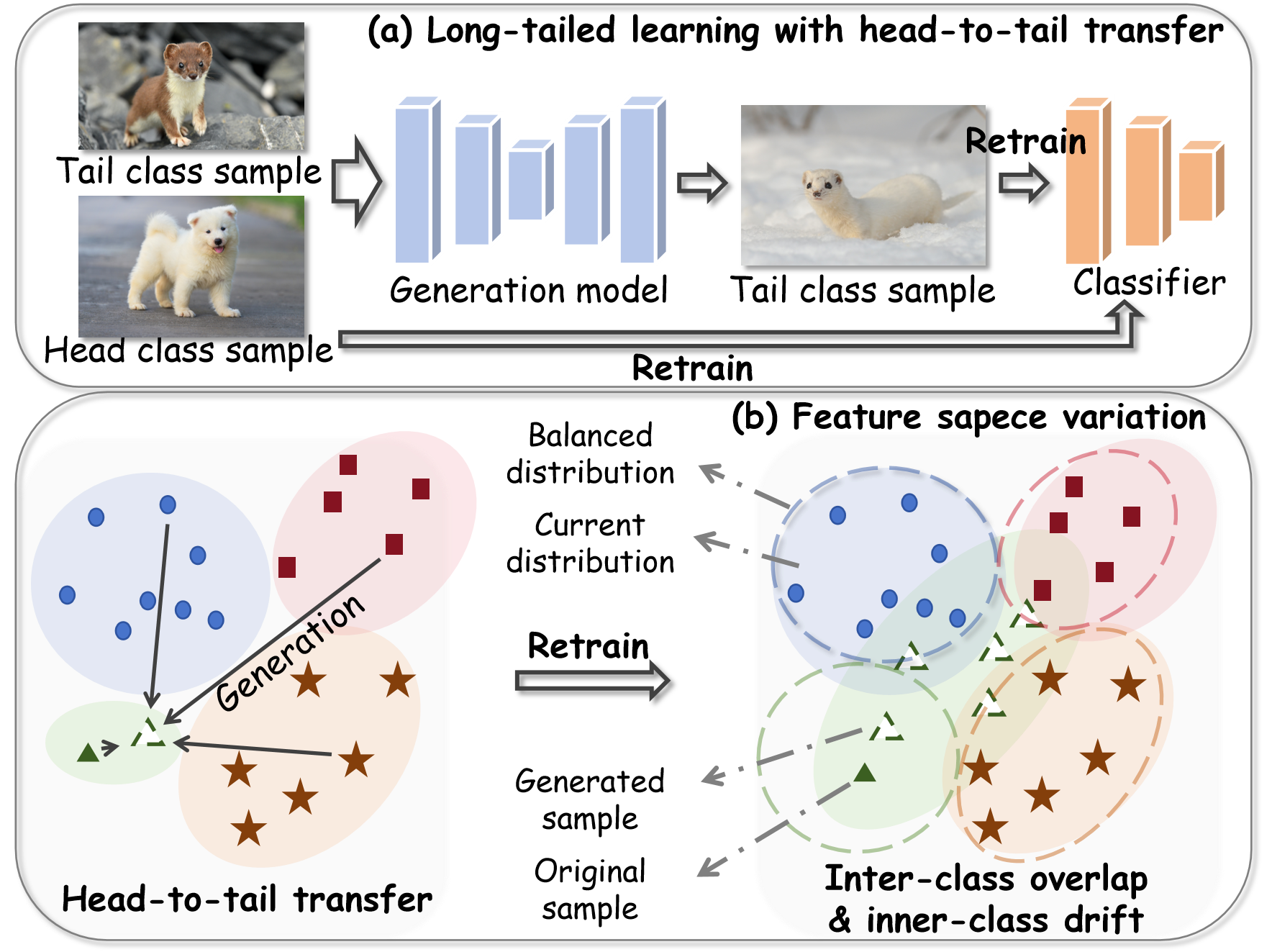}}
\vspace{-15pt}
\caption{Pipeline of head–to–tail transfer for tail sample generation, resulting in inter-class overlap, intra-class drift, and blurred decision boundaries.}
\vskip -15pt
\label{fig:intro}
\end{figure}

From a different perspective, augmentation-based approaches~\cite{long2022retrieval, qin2023class} address data scarcity by supplementing training sets with additional samples. Among them, generative modeling offers robust performance by exploiting data priors. Recent works employ powerful diffusion models~\cite{han2024latent} along with head-to-tail transfer to mitigate generator bias and promote a more uniform decision space~\cite{shi_how_2023, shao2024diffult}. These methods effectively leverages limited training data to improve the quality of generated samples and mitigate the generator’s bias toward head classes as shown in~\autoref{fig:intro} (a). However, while prior methods pursue uniform decision space, they rarely analyze the boundary blurring and tail drift induced by head–to–tail transfer~\cite{kim2020m2m, liu_lcreg_2024}. As shown in~\autoref{fig:intro} (b), head-to-tail transfer helps balance the decision space but inevitably induces latent, uncontrolled nonlocal feature leakage that entangles class distributions, leading to inter-class overlap and shifts in tail class decision boundaries. Without a balanced reference, the leakage is difficult to disentangle, leaving a “more uniform yet highly overlapped” decision space and limiting gains in tail-class learning. Specially, we term this problem boundary ambiguity.

%基于视觉的深度学习方法的出现给人们的日常生活带来极大的便利和效率的提升，作用于身份验证、物体检测、视频处理等各个领域。随之而来的，各个领域的图像训练数据也变得越来越重要，甚至被称为数字黄金【引数字黄金的论文】。数据的质量和数量极大的影响着基于视觉的深度学习方法的效果【引大模型数据量变化研究相关的论文】。然而，由于数据本身的分布不平衡和人工采集的限制，数据集大多存在长尾现象——即少数类别占据大部分样本【引长尾分布论文】。这极大程度的造成了分布空间上少样本类(尾部类)被多样本类(头部类)压迫导致精度下降，最终导致深度学习模型整体表现变差，难以处理少见但重要的情况【引两篇应用级论文吧】。
%因为这个原因，许多研究者尝试从特征提取和训练流程入手，缓解由于数据分布不平衡导致的模型对尾部类表征不足和识别精度差的问题。如XXX【】引入软约束，扩展尾部类特征空间以减轻特征空间上头部类对尾部类的压迫。XXX【】提出了XX损失，增强尾部类对模型训练的贡献，以减轻模型对头部类的偏向。然而，这些方法始终无法解决尾部类特征缺失这个根本问题。因此，研究者们尝试补全缺失的尾部类数据。
To make the boundary ambiguity measurable under generation-based long-tailed learning, we design three tailored metrics: (i) inter-class overlap degree which quantifies pairwise overlap among class distributions; (ii) tail outlier rate for estimating the class decision boundary shift; and (iii) generation confidence for intuitively showing the feature entanglement. When applied across generated datasets, these metrics consistently reveal blurred decision boundaries and tail distribution drift under head–to–tail transfer. These findings caution that \textbf{indiscriminate use of head class features to expand the tail class can lead to decision boundary damage}, expansion based on head-to-tail transfer is therefore a double-edged sword. 

%补全缺失的数据集的方法可以分为三种，分别是引入外部数据，数据增强和数据生成。经典的引入外部数据的方法，如XXX，通过网络搜索和筛选补充缺失的尾部类数据。然而这样引入的数据不仅难以拟合原数据集分布【这有没有论文呀】，而且容易将模型暴露在后门攻击中【diffusionLT】。而数据增强的方法随机性太强，难以提供足够的缺失信息，错误的数据增强甚至会进一步降低模型的整体精度【类数据增强学习】。与前两种方法不同的是，数据生成的方法仅用已有数据生成新的补充数据集，这不仅规避了将模型暴露在后门攻击中的危险，还利用了已有数据集的信息避免补充数据集与原数据集不拟合。最新的数据生成方法，用已有数据训练一个强大的diffusion模型，用于补充缺失尾部类数据【引两篇论文】。为了减轻由于数据集分布不平衡导致数据生成网络偏向头部类，部分研究者尝试用头部类数据辅助尾部类数据生成。如，XXX【】用mixup的方法，混合头部类背景和尾部类前景创建新的尾部类样本。XXX【】在训练diffusion模型时，混合头尾类同步训练，以为尾部类训练提供额外特征信息。然而，如【图1】所示，这种头尾混淆策略容易混淆类间分类边界，造成特征分布空间重叠，尤其是当引入的头部类与当前尾部类特征距离较远时，详情见【section2】。

Motivated by these observations, we introduce the \textbf{G}enerative \textbf{B}oundary-aware \textbf{G}eneration (\textbf{DBG}) framework to address the boundary ambiguity for promoting tail class learning. DBG integrates a boundary-aware generator with a classifier-driven bifurcated cleaning pipeline to generate informative near-boundary samples, thereby repopulating and refining the decision boundaries. Specifically, the boundary-aware generator introduces adversarial declassification noise to suppress dominant class-specific patterns and reconstruct each sample toward its $k$-nearest alternative classes, enriching boundary features for long-tailed learning. Furthermore, to mitigate hard samples arising from long-tail-biased labeling and generator–classifier space mismatch, the classifier-driven bifurcated cleaning pipeline integrates class-wise prototype-distance and confidence–credibility filtering to identify and discard harmful samples. Overall, DBG effectively addresses the scarcity of tail class data and sharpens the decision boundaries, reconstructing a "more uniform and separable" decision space that substantially improves tail class precision.
%为了解决分类边界混淆问题，受到【】和对抗性训练方法的启发，我们提出了用对抗攻击的思想，生成位于分类边界的样本，用于帮助模型细化分类边界，减轻头尾特征混淆的情况，从而达到更好的识别精度。具体的，我们修改了用diffusion生成新样本的过程，不再从噪声出发随机生成目标类样本，而是引入对抗性攻击的去类信息思想，通过加噪过程将源数据集样本的原始类信息去掉，然后再通过有目标的去噪过程，重新将加噪图片去噪为近似类图像。然而，由于这种对抗性攻击流程存在不稳定性，且位于分类边界的样本难分，难以给攻击失败样本打上正确的标签。因此，我们通过样本-类原型距离筛选掉部分偏移过远样本，并用置信度分析剔除攻击特征偏移过大样本。

%我们的贡献如下：
%发现并设立了三个指标观测头尾特征混淆机制带来的特征空间重叠和分类边界混淆问题
%提出了一个数据生成方法，细化特征分类空间
%进一步提出筛选机制，去除无用样本
%实验效果？
Our main contributions are summarized as follows:
\begin{itemize}
\item To our knowledge, we present the first systematic analysis of boundary ambiguity induced by head-to-tail transfer, using three metrics: inter-class overlap degree, outlier rate, and generation confidence.
\item We propose a Decision Boundary-aware Generation framework to generate informative samples for enriching near-boundary features and sharpening decision boundaries in long-tailed learning.
\item Extensive experiments demonstrate that the proposed framework reduces inter-class overlap and improves long-tailed learning performance for all generative baselines, demonstrating its effectiveness and generalization ability.
\end{itemize}
\section{Related Works}
\label{sec:Related works}
\subsection{Augmentation-based Long-tailed Learning}
Data quantity and quality critically determine the performance of deep learning classifiers, while real-world datasets are often long-tailed~\cite{andersonLongTailWhy2006}. Prior work introduces additional samples to rebalance long-tailed datasets for improving overall classify accuracy~\cite{ahn_cuda_2023, zhao2025learning, zhang2024long, lt2026hong}. However, such strategies can introduce out-of-distribution features and even backdoor risks, degrading both accuracy and security of the classifiers. In response, generator-based approaches emerged. For instance,~\citet{liu_lcreg_2024} generates tail samples by mixing head-shared and tail-specific features.~\citet{qin2023class} leverages diffusion model~\cite{rombach2022high} trained with the long-tailed data to generate missing samples. Furthermore,~\citet{shao2024diffult} combines diffusion with head-to-tail transfer strategy to mitigate head class bias. In this paper, we introduce generation based on adversarial attack to reduce inter-class overlap and restore blurred decision boundaries for long-tailed learning.
%数据的数量和质量极大的影响着基于深度学习的分类模型的有效性。然而，由于数据集天然的分布不平衡性和人工采集的限制，现实数据往往呈现长尾分布【】。这使得训练完好的分类模型难以正确分类数据量稀少的尾部类样本，且在特征空间上呈现头部类分布压缩尾部类分布的现象。为了解决该问题，研究者们尝试引入或生成额外的样本以重平衡呈现长尾分布的不平衡数据集。如，XXX【】采用定向随机增强的方法补充不同类样本。进一步的，为了引入更多样的特征以丰富尾部类，XXX【】引入打上伪标签后的外部数据重平衡训练数据集。然而，这样容易引入特征分布外的数据和后门攻击，导致识别精度和模型安全性下降。为了解决该问题，基于原始数据集重建和生成器训练的方法应运而生。如，XXX【】混合头部类通用特征和尾部类类特有特征以生成新的尾部类样。进一步的，XXX【】引入热门的图像生成网络——diffusion【】，通过用原始数据集重训练，以达到生成原始特征域内的缺失数据。而XXX【】结合类间特征混合（也就是head to tail）的思想，在训练diffusion生成时混合部分头部类和尾部类样本以提供更多训练信息，从而减轻diffsuion模型对头部类的偏向，以生成更有效的尾部类样本。尽管这些方法取得了较好生成效果，我们发现这种头尾混淆的策略容易导致类间分布重叠度增高，分类边界混淆。为了解决该问题，我们提出引入对抗性攻击思想，生成接近分类边界的额外样本，以细化分类边界，增强特征空间可分性。
\subsection{Adversarial Samples}
Adversarial samples are inputs perturbed with small, human-imperceptible noise that flip the prediction of classifier, while preserving visual semantics~\cite{goodfellow2014explaining, dong2018boosting, yang2024decoupling}. Beyond robustness evaluation, they also serve as data augmentation to improve generalization~\cite{goodfellow2014explaining, zhao2022adversarial}. In long-tailed learning they have been used to mitigate imbalance data distribution by reconstructing tail samples via adversarial attacks toward head samples~\cite{kim2020m2m}. In this work, we generate near-boundary adversarial samples to offer additional boundary knowledge for restoring decision regions and sharpen separability.

%对抗性攻击是一种通过给图像加上人眼无法察觉的微小扰动，改变图像原始类信息，使得分类网络分类错误的方法【初始对抗性攻击】。它可以细分为有目标的对抗性攻击和无目标的对抗性攻击。如XXX【】引入动量缩放持续的为图像添加微小扰动，以达到最终使图像被分类为目标类的目的。而XXX【】通过多步计算样本当前与最短分类边界方向，实现抹除原始类信息的目标。随着基于Unet【】的图像生成方法的发展，也出现了一些利用深度特征提取backbone提取深层图像特征以重建视觉上相似但无法正确分类的对抗性样本的方法。如XXX【】引入替代分类器指导训练基于Unet的对抗性样本生成器。进一步的，XXX【】设计了基于GAN【】的对抗性样本生成器。这些生成的对抗性样本除了可以应用于检测分类网络的鲁棒性外，还可以加入训练集重训练，以达到提高分类器泛化能力和鲁棒性的目的【初始对抗性训练】。在解决长尾学习问题中，对抗性攻击也被作为一种扩充尾部类样本的手段。如XXX【M2M】通过对抗性攻击改变头部类样本的类别信息，以为尾部类训练提供额外特征。在本文，我们引入对抗性攻击生成接近分类边界的额外样本，在补充尾部类样本的同时，构建更易分的特征空间。
\section{Analysis of Boundary Ambiguity}
\label{sec:observation}
The uniformity and separability of the feature space largely determine the difficulty of classification. Owing to severe class imbalance, long-tailed datasets often yield post-training feature spaces with uneven inter-class density. Existing head-to-tail training strategies mainly aim to equalize inter-class distributions while paying less attention to improving inter-class separability. To further observe this, in this section, we introduce three complementary metrics to jointly quantify the uniformity and separability of the learned feature space, including inter-class overlap degree, outlier sample rate, and generation confidence.

\subsection{Boundary Ambiguity Measurement Metrics}

%特征空间分布的均匀性和可分性极大的影响着分类模型的分类难度。长尾数据集由于其极端不平衡分布，导致模型训练后得到的特征空间往往存在类间分布不均，且尾部类特征空间被侵占的特性。头转尾的训练策略仅考虑了如何使得类间特征分布变得均匀，并没有考虑如何使得类间特征空间更可分。在本文中，我们分别设计了类间重合度，离群点（噪点）和生成样本置信度三个指标来量化模型的特征空间分布均匀性和可分性。

\paragraph{Inter-class overlap degree.}
Our goal is to quantify how an generated training set reshapes the separability of the learned feature space. Moreover, since the classification results are directly determined by the test set, we restrict our analysis to the changes in the feature space on the test data.
To this end, the inter-class overlap degree metric is defined based on von Mises–Fisher (vMF) hyperspherical projection and the Bhattacharyya coefficient (BC) to measure the overlap degree between each pair of classes.

Specifically, test features are first extracted by a well-trained encoder and then projected onto the unit hypersphere via $L2$ normalization by $y=\frac{z}{\|z\|_2}$, which enables uniform comparability of inter-class overlap. $z \in\; \mathbb{S}^{d-1}$ denotes the extracted feature and $y$ is the normalized feature after projection.
After that, for each class $c\in\{1,\dots,K\}$ with normalized test features $\{y_i^{(c)}\}_{i=1}^{n_c}\subset\mathbb{S}^{d-1}$, we fit a vMF density for representing intra-class distribution:
\begin{equation}
\label{eq:fit}
\begin{aligned}
p_c(y) \;=\; C_d(\kappa_c)\,\exp\!\big(\kappa_c\,\mu_c^\top y\big),
\end{aligned}
\end{equation}
where $\mu_c\in\mathbb{S}^{d-1}$ is the mean direction and $\kappa_c\!\ge\!0$ is the concentration, both estimated by maximum likelihood from $\{y_i^{(c)}\}$. $C_d(\cdot)$ is the vMF normalizer. 
Then, pairwise inter-class overlap is measured by BC as follows.
\begin{equation}
\label{eq:logBC}
\widehat{\mathrm{BC}}(c,c')
\;=\;
\log\!\left(
\frac{1}{m}\sum_{i=1}^{m}
\sqrt{\,p_c(y)\,p_{c'}(y)\,}
\right).
\end{equation}
Lower $\widehat{\mathrm{BC}}(c,c')$ indicates a more separable feature space between $c$ and $c'$.

\begin{figure}
\centering
{\includegraphics[width=0.47\textwidth]{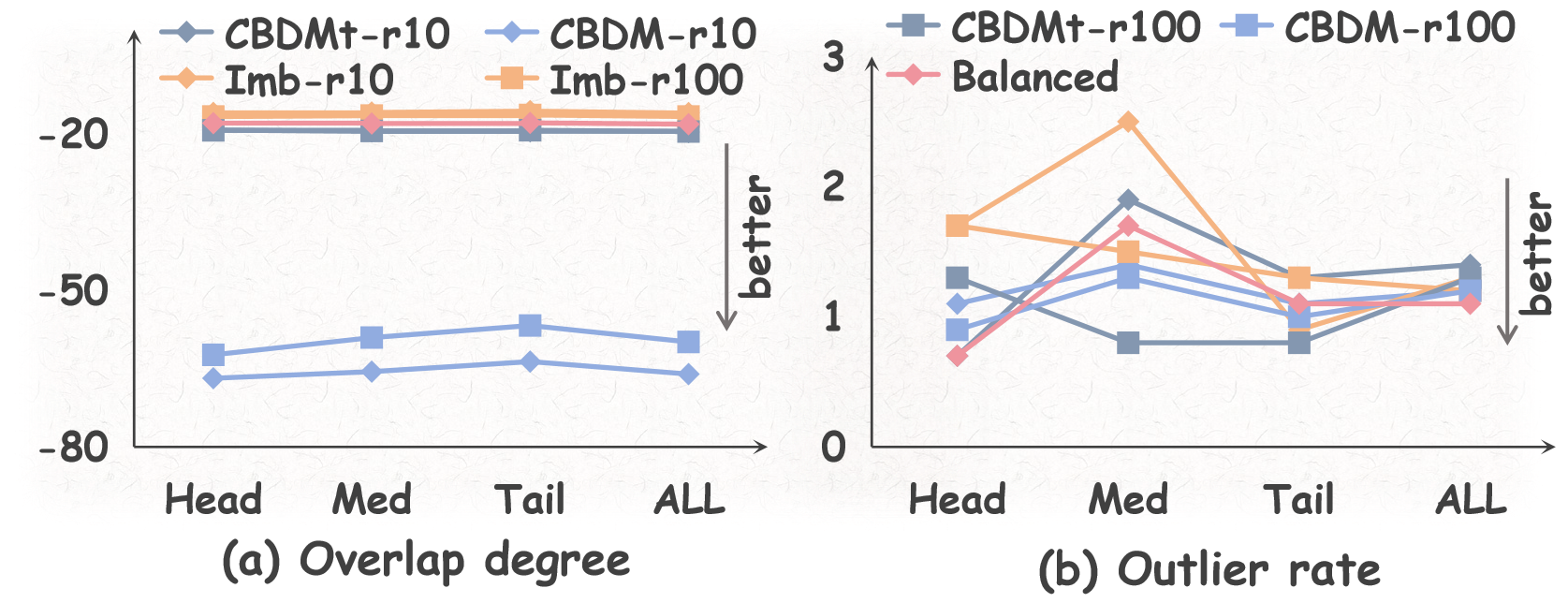}}
\vspace{-12pt}
\caption{(a) The results of inter-class overlap degrees in the feature spaces learned by four different classifiers. (b) The results of Outliers sample rate. Imb-r represents the ResNet-32 trained on the imbalanced dataset with imbalanced ratio set to r, and CBDMt denotes the classifier trained on the CBDM-based~\cite{qin2023class} dataset with head-to-tail transfer.}
% \emph{(i) balanced dataset, 
% (ii) long-tailed dataset, 
% (iii) extended dataset generated by class-balancing diffusion models (CBDM-based)}~\cite{qin2023class}\emph{, and
% (iv) extended dataset generated by CBDM-based along with head-to-tail transfer.} 
\vskip -20pt
\label{fig:t1}
\end{figure}

To investigate the impact of imbalance ratio and head-to-tail transfer on the decision space of the classifier, inter-class overlap degree is measured under four varint datasets.
% \emph{(i) balanced dataset, 
% (ii) long-tailed dataset, 
% (iii) extended dataset generated by class-balancing diffusion models (CBDM-based)}~\cite{qin2023class}\emph{, and
% (iv) extended dataset generated by CBDM-based along with head-to-tail transfer.} 
As shown in \autoref{fig:t1} (a), the imbalanced setting yields higher inter-class overlap, indicating poorer separability. 
CBDM-based reduces inter-class overlap by increasing feature diversity via diffusion-based generation. 
However, introducing head-to-tail transfer increases overlap again, proving our hypothesis that coarse-grained head-to-tail transfer can harm feature-space separability.

\begin{figure}
\centering
{\includegraphics[width=0.47\textwidth]{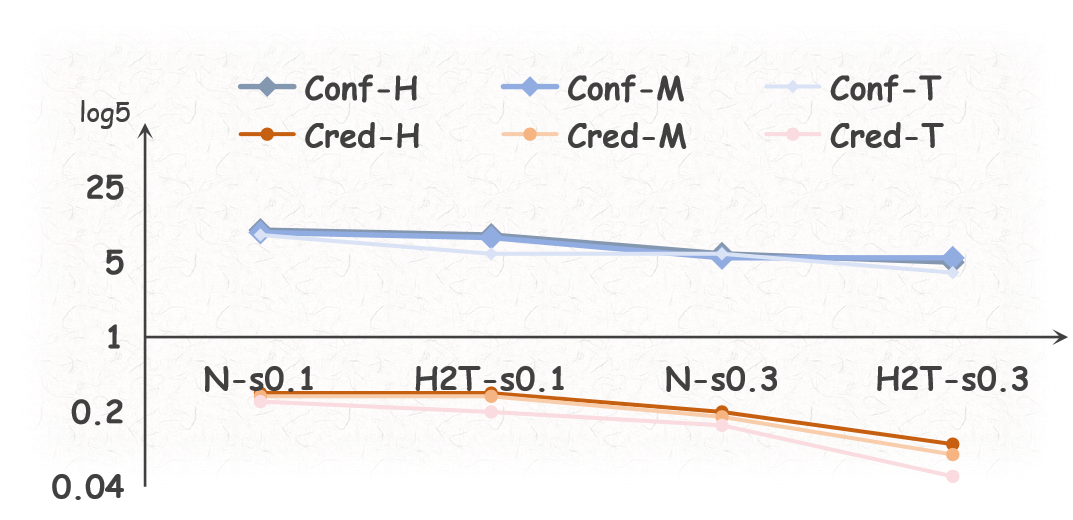}}
\vspace{-12pt}
\caption{The results of generation confidence with different $s$ on balanced classifier. N represents normal training, while H2T represents trained with head-to-tail transfer. H, M, and T represent statistics based on head, medium, and tail classes, respectively.}
\vskip -15pt
\label{fig:t2}
\end{figure}
%类间重合度指标旨在计算经过扩展后的训练集训练后的分类模型的特征空间的在测试集上的类间重叠程度，以近似的展示特征空间的可分性，量化扩展数据集对模型特征空间的影响。具体的，所有类的特征都先被投影至同一个vmf球形特征空间，以实现对所有类的统一特征对比计量，如下：
%公式
%进一步的，bathery距离算法被引入以量化类间重合度。
%我们分别使用了平衡的cifar100数据集，不平衡ratio为100的cifar100数据集，经典的diffusion长尾样本生成方法——OCLT_none扩展后的cifar100数据集和其改进版增加了头尾混合训练机制的OCLT方法扩展后的cifar100数据集训练resnet分类网络并进行计算类间重合度，结果如图2所示。可以看出不平衡cifar-100数据集训练的分类器类间特征重合度更高，而OCLT_none方法通过diffusion生成更多样的特征扩展了各类的特征量，从而达到了减小类间重合度的效果。而OCLT方法由于引入了头尾混合训练策略，导致类间重合度进一步上升。由此，应证了我们所提出的粗粒度的头转尾策略会影响特征空间可分性的猜想。
\paragraph{Outlier sample rate.}
Due to the instability of parameter changes during model training, it is very difficult to directly evaluate whether the generated dataset exhibits feature space shift. But ideally, the feature space learned by the model should be able to uniformly project the test set samples onto their corresponding class distributions. Based on this, we measure the degree of intra class shift by testing the approximate distribution of sample features in the feature space. Specifically, to verify whether the head-to-tail mechanism induces unfavorable intra-class distribution shift, we flag test samples that deviate excessively from their class-wise neighborhoods in the learned feature space.
Given a trained encoder, the within-class nearest-neighbor distance is computed for each sample:
\begin{equation}
\label{eq:distance}
\begin{aligned}
d_i^{(c)} \;=\; \min_{\substack{j\neq i\\ 1\le j\le n_c}} \,\big\|\,z_i^{(c)}-z_j^{(c)}\big\|_2,
\end{aligned}
\end{equation}
where $z_i^{(c)}$ is the $i$-th feature of class $c$ and $n_c$ is the number of test samples in class $c$.  
$d_i^{(c)}$ is the class-$c$ nearest-neighbor distance of sample $i$. Let $\bar d_c=\tfrac{1}{n_c}\sum_{i=1}^{n_c} d_i^{(c)}$ and
$s_c=\operatorname{std}\!\big(\{d_i^{(c)}\}_{i=1}^{n_c}\big)$,
we standardize nearest-neighbor distances and mark outliers by a threshold $\lambda$ as follows.
\begin{equation}
\label{eq:mark}
\begin{aligned}
s_i^{(c)} \;=\; \frac{\big|d_i^{(c)}-\bar d_c\big|}{s_c},\qquad
o_i^{(c)} \;=\; \mathbb{I}\!\big[s_i^{(c)}>\lambda\big],
\end{aligned}
\end{equation} 
where $\bar d_c$ is the mean of $\{d_i^{(c)}\}_{i=1}^{n_c}$. 
$\operatorname{std}(\cdot)$ denotes the standard deviation so that $s_c$ is the standard deviation of the nearest-neighbor distances in class $c$. $\mathbb{I}[\cdot]$ is the judgement function (1 if the predicate is true, 0 otherwise). 
$o_i^{(c)}\in\{0,1\}$ is the outlier label. Finally, the outlier sample rate $\eta_c$ can be defined as $
\eta_c
=
\frac{1}{n_c}
\sum_{i=1}^{n_c}
o_i^{(c)}
$ to measure the distribution shift. 

% \begin{table}[]
% \caption{Noise ratio.}
% \centering
% \label{tab:my-table}
% \begin{tabular}{c|c|cccc}
% \hline
% \rule{0pt}{2.8ex}
% \textbf{Trainset} & \textbf{Ratio} & \textbf{Head} & \textbf{Med} & \textbf{Tail} & \textbf{All} \\[0.6ex] \hline
% \rule{0pt}{2.3ex}
% Balanced & 1   & \textbf{2.0} & 2.3 & \textbf{1.7} & \textbf{2.03} \\
% Imb      & 10  & 2.5 & 2.3 & 2.0 & 2.28 \\
% OCLT     & 10  & 2.1 & 2.0 & 2.4 & 2.18 \\
% CBDM     & 10  & 2.3 & \textbf{1.9} & 2.2 & 2.15 \\ \hline
% \rule{0pt}{2.3ex}
% Imb      & 100 & 2.0 & 1.6 & 2.5 & 2.00 \\
% OCLT     & 100 & \textbf{1.8} & 2.0 & 2.1 & 1.97 \\
% CBDM     & 100 & 2.0 & \textbf{1.9} & \textbf{1.9} & \textbf{1.92} \\ \hline
% \end{tabular}
% \end{table}
% \begin{figure}
% \centering
% {\includegraphics[width=0.45\textwidth]{LaTeXAuthor_Guidelines_for_Proceedings/fig/figt2-1.png}}
% \vspace{-10pt}
% \caption{The results of Outliers sample ratio in the feature spaces learned by four different classifiers.}
% \vskip -10pt
% \label{fig:t2}
% \end{figure}
Similarly, we evaluate outlier sample rate on classifiers trained on four variant datasets. As shown in~\autoref{fig:t1} (b), increasing imbalance ratio of the dataset and introducing head-to-tail transfer both lead to more test-time outliers in the feature space, especially for tail classes. This indicates head-to-tail transfer leads to tail class distribution shift, degrading tail class accuracy. It is also worth-noted that higher imbalanced rate can increase the model bias toward head classes, resulting in less outlier samples.

%离群点指标从测试集是否会出现类偏移较大的特征点的角度，侧面验证加入头转尾机制后类特征空间是否发生了不利偏移，以证明特征空间变得更为混乱。具体的，我们首先计算类内特征点平均距离，然后以此找出极端偏移的离群点，如下所示：
%公式
%同样的，我们在四个不同的训练集训练的分类器进行验证，结果如图3所示。可以看出随着训练集的不平衡程度和头尾特征混合策略的使用，在特征空间内测试集的离群点变多，尤其是对尾部类。
\begin{figure*}[htbp]
\centering
\includegraphics[width=1\textwidth]{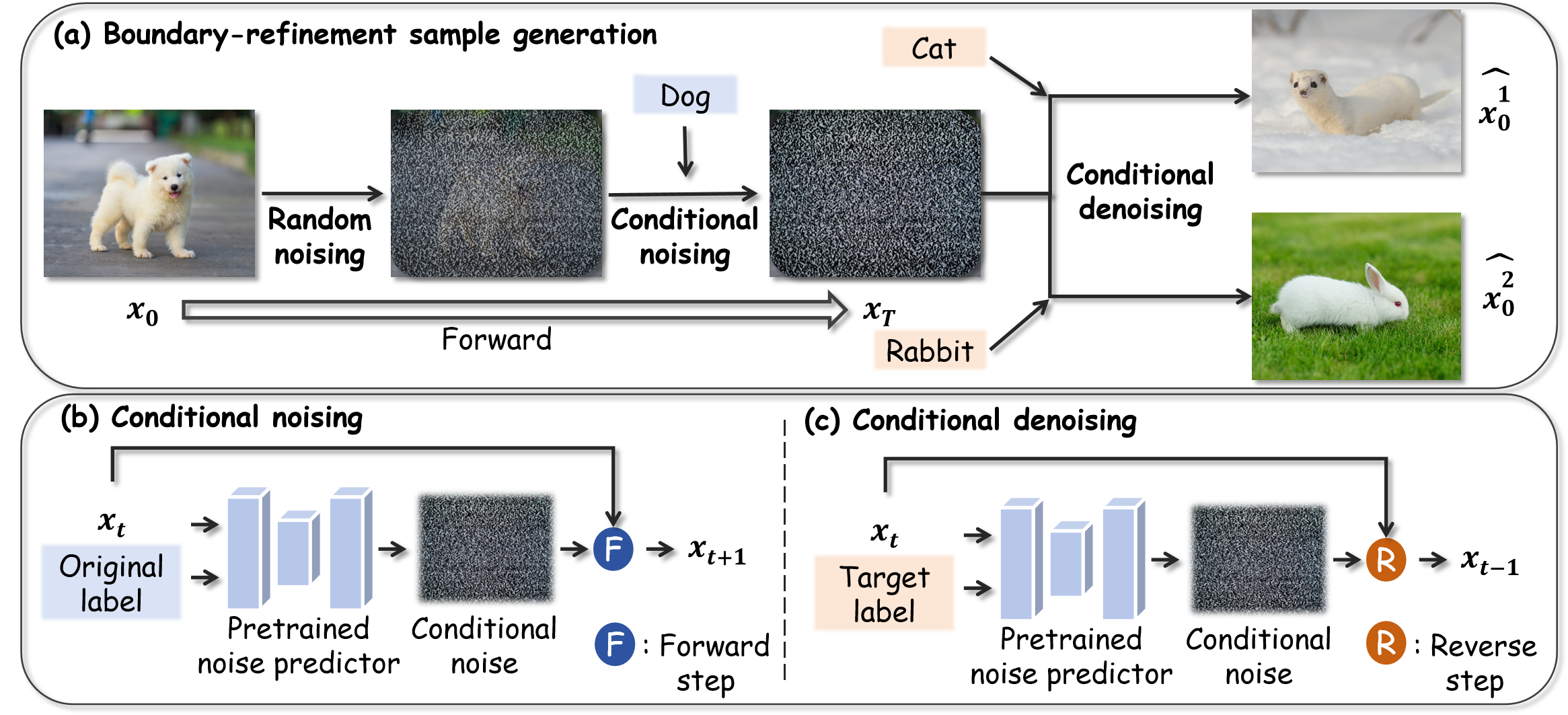}
\vspace{-20pt}
\caption{The basic pipeline of boundary-aware sample generator, including (b) conditional noising and (c) conditional denoising stages.}
\label{fig:gen}
\vskip -4pt
\end{figure*}

% \usepackage{multirow}
% \begin{table}[]
% \centering
% \caption{The results of generation confidence with different $s$ on balanced classifier.}
% \label{tab:conf}
% \setlength{\tabcolsep}{4pt}
% \begin{tabular}{cc|ccc|ccc}
% \hline
% \rule{0pt}{2.3ex}
% \multirow{2}{*}{H2T} & \multirow{2}{*}{$s$} & \multicolumn{3}{c|}{Confidence} & \multicolumn{3}{c}{Credibility} \\ 
%   &     & Head & Med  & Tail & Head & Med  & Tail \\ \hline \rule{0pt}{2.3ex}
%   & 0.1 &\textbf{9.60} & \textbf{9.68} & \textbf{8.92} & 0.32 & \textbf{0.31} & \textbf{0.29} \\
% \checkmark & 0.1 & 9.24 & 9.39 & 6.40 & \textbf{0.37} & \textbf{0.31} & 0.17 \\
%   & 0.3 & 6.20 & 6.44 & 5.79 & 0.16 & 0.13 & 0.11 \\
% \checkmark & 0.3 & 5.34 & 5.14 & 4.36 & 0.13 & 0.08 & 0.06 \\ \hline
% \end{tabular}
% \end{table}

\paragraph{Generation confidence.} To intuitively quantify boundary ambiguity induced by head–to–tail transfer on diffusion models, generation confidence is defined as the class-wise average prediction confidence of generated data. Specifically,
a diffusion probabilistic model consists of a forward noising process and a reverse denoising process. The generation process usually begins with a random Gaussian noise sample and, by predicting and removing the noise iteratively follows standard Classifier-free guidance (CFG)~\cite{ho2021classifier, ho2020denoising}:
\begin{equation}
\label{eq:reverse-guided}
{\boldsymbol{\hat{\varepsilon}}_{t}}
= (1-s)\cdot\boldsymbol{\varepsilon}_{\theta}(\mathbf{x}_t,\varnothing,t)+s\cdot\boldsymbol{\varepsilon}_{\theta}(\mathbf{x}_t,\mathbf{c},t),
\end{equation}
where \(\boldsymbol{\varepsilon}_{\theta}(\mathbf{x}_t,\mathbf{c},t)\) and
\(\boldsymbol{\varepsilon}_{\theta}(\mathbf{x}_t,\varnothing,t)\) denote the
conditional and unconditional noise predictions, respectively. $s$ is the guidance scale.
To directly validate how head-to-tail confusion affects the feature space learned by diffusion models, we form a modified CFG guidance for a target class $y_t\in\mathcal{C}_{\text{tar}}$ and a disturbing class $y_d\in\mathcal{C}_{\text{dis}}$ as follows.
\begin{equation}
\label{eq:modified-reverse-guided}
{\boldsymbol{\hat\varepsilon}}_{tm}
= (1-s)\cdot\boldsymbol{\varepsilon}_{\theta}(\mathbf{x}_t,y_t,t)
   + s\cdot\boldsymbol{\varepsilon}_{\theta}(\mathbf{x}_t, y_d,t).
\end{equation}

The generated samples under modified guidance are then feed into a ResNet classifier $f_\phi$ trained on the balanced dataset to obtain logits $\ell(x_g)\in\mathbb{R}^K$ and probabilities $p(x_g)=\operatorname{softmax}(\ell(x_g))$.
The confidence and credibility of sample $x_g$ is then computed to measure the generation confidence as follows.
\begin{equation}
\label{eq:confidence}
\begin{aligned}
 Conf(x_g) \;&=\; \ell_{y_t}(x_g),\\
Cred(x_g) \;&=\; p_{(1)}(x_g)\;-\;p_{(d)}(x_g),
\end{aligned}
\end{equation}
where $\ell_{y_t}(x_g)$ denotes the $y_t$-th logit of $\ell(x_g)\in\mathbb{R}^K$. $p_{(1)}(x_g)=\max_{c} p_c(x_g)$ and $p_{(d)}(x_g)$ is the predicted probability of $x_g$ from disturbing class. A lower confidence and a lower credibility indicates stronger boundary ambiguity. 

As shown in \autoref{fig:t2}, after employing head-to-tail transfer, the prediction confidence of the generated samples reduce, indicating that this strategy makes head-specific features leak into tail class generated samples, leading to boundary ambiguity. It is worth noted that the prediction credibility also reduce, indicating that the generator also occurs boundary ambiguity for disturbance classes.
% \begin{figure}
% \centering

% % 第一行两张图
% \begin{subfigure}{0.23\textwidth}
%  \centering
% \includegraphics[width=\linewidth]{LaTeXAuthor_Guidelines_for_Proceedings/fig/100_001_1.png}
%     \caption{Balanced classifier}
%     \label{fig:sub1}
% \end{subfigure}
% \hfill
% \begin{subfigure}{0.23\textwidth}
% \centering
% \includegraphics[width=\linewidth]{LaTeXAuthor_Guidelines_for_Proceedings/fig/100_001_001.png}
%     \caption{Unbalanced classifier}
%     \label{fig:sub2}
% \end{subfigure}

% \vspace{0.5em} % 调整上下间距

% % 第二行两张图
% \begin{subfigure}{0.23\textwidth}
% \centering
% \includegraphics[width=\linewidth]{LaTeXAuthor_Guidelines_for_Proceedings/fig/100_001_d.png}
%     \caption{(OCLT classifier}
%     \label{fig:sub3}
% \end{subfigure}
% \hfill
% \begin{subfigure}{0.23\textwidth}
% \centering
% \includegraphics[width=\linewidth]{LaTeXAuthor_Guidelines_for_Proceedings/fig/100_001_dn.png}
%     \caption{CDBM classifier}
%     \label{fig:sub4}
% \end{subfigure}

% \caption{Inter-class overlap.}
% \label{fig:noise}
% \end{figure}

%为了更直接的验证头尾混淆策略对特征空间的影响，我们通过调整采样过程中的cfg条件来展示头转尾策略对diffusion样本生成质量的影响。具体的，采样时cfg生成条件做出如下调整：
%公式
%接下来，生成的样本被输入用平衡cifar数据集训练的分类器，以得到生成样本置信度和可信度，结果如图4所示。其中，置信度以样本目标类的logit表示，而可信度以样本最高类和次高类置信度表示，表示如下：
%公式
%可以看出，由于在训练过程中，混合了头部类和尾部类的特征，导致模型生成的样本容易引入其他类特定信息。

\section{Decision Boundary-aware Generation}
\label{sec:method}
The DBG comprises two components: (i) a boundary-aware sample generator that generates samples near the decision boundary to restore sharper decision boundaries; and (ii) a classifier-driven bifurcated data-cleaning pipeline that filters hard and drifted samples, preventing secondary distortion of the feature space caused by long-tailed bias.
%我们提出的对抗性边界强化框架有两个组件，包括边缘重细化样本生成组件用于生成靠近分类边界的样本以帮助模型恢复更清晰的分类边界，和难分样本清理组件用于筛去无用和难分样本以避免因无法正确分类而对特征空间造成二次害。

\begin{figure*}[htbp]
\centering
{\includegraphics[width=1\textwidth]{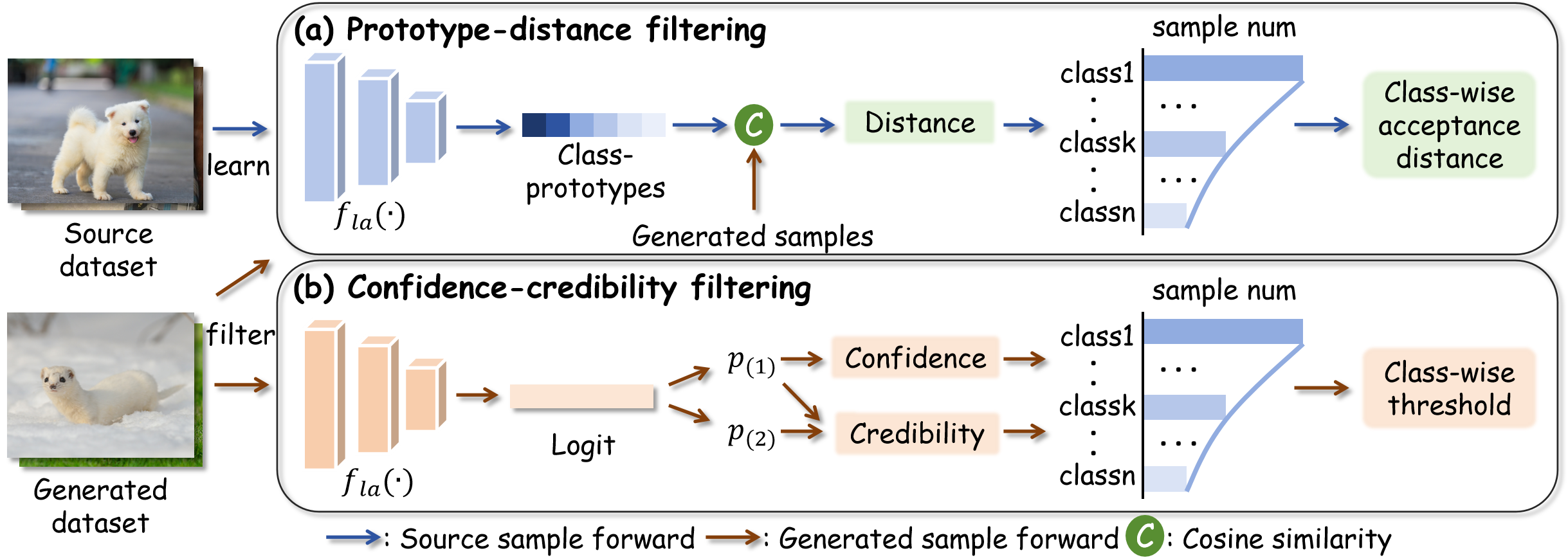}}
\vspace{-20pt}
\caption{The workflow of classifier-driven bifurcated data-cleaning, consists of prototype-distance and confidence-credibility filtering.}
\vskip -10pt
\label{fig:filter}
\end{figure*}

\subsection{Boundary-aware Sample Generator}
The boundary-aware sample generator consists of two stages, including conditional noising to erase class-specific features for pulling the samples to its decision boundary, and conditional denoising for assigning target labels to the reconstructed samples, as shown in~\autoref{fig:gen}.

For conditional noising, with diffusion trained with horizon \(T\), schedule \(\beta_t\in(0,1)\), and \(\alpha_t=1-\beta_t\), \(\bar{\alpha}_t=\prod_{s=1}^{t}\alpha_s\), the input sample $x_0$ is first perturbed by random noise $\hat{\epsilon}_r$ to push it into a middle timestep for easier modifying deep class-specific features~\cite{mengsdedit}, as follows.
\begin{equation}
x_{m} \;=\; \sqrt{\bar{\alpha}_{m}}\,{x}_0 \;+\; \sqrt{1-\bar{\alpha}_{m}}\,\hat{\epsilon}_r,
\label{eq:forward_closed_form_m}
\end{equation}
where $m=T/2$ and $\hat\epsilon_r$ denotes standard Gaussian noise.

After that, to suppress the class-specific features, $x_m$ is forwarded under the guidance of source label \(y_{\text{sl}}\) following~\autoref{eq:forward_closed_form_m} iteratively and the target class-specific noise $\boldsymbol{\hat\varepsilon}_c$ in it is obtained via the noise predictor \(\boldsymbol{\varepsilon}_\theta(\cdot)\) as follows.
\begin{align}
\boldsymbol{\hat{\varepsilon}}_c &= \boldsymbol{\varepsilon}_\theta\!\big(x_t,\,  y_{\text{sl}},\, t\big),
\label{eq:noise_pred_src}
\end{align}
where $x_t$ denotes the current noisy sample.
Specifically, the noising iteration update for \(K=T/10\) steps starting at $m$. Which prevents the reconstructed sample from drifting too far from the source manifold while effectively erasing class-specific features to produce stable near-boundary variants.

For conditional denoising, we perform reverse denoising with standard CFG (following~\autoref{eq:reverse-guided}) as follows.
\begin{equation}
\label{eq:cfg_whole}
\begin{aligned}
\hat{x}_0
=&\; \frac{x_t - \sqrt{1-\bar{\alpha}_t}\,\hat{\boldsymbol{\varepsilon}}_{t}(x_t,t,\tilde{y})}{\sqrt{\bar{\alpha}_t}},\\
x_{t-1}
=&\; \sqrt{\bar{\alpha}_{t-1}}\,\hat{x}_0 \;+\; \sqrt{1-\bar{\alpha}_{t-1}}\,\hat{\boldsymbol{\varepsilon}}_{t}(x_t,t,\tilde{y}),
\end{aligned}
\end{equation}
where \(\tilde{y}\) denotes the label close to $x_0$.
By iteratively removing predicted noise, the generated samples remain near the decision boundary and avoid large semantic jumps while providing additional knowledge of \(\tilde{y}\) for the classifier.

Specifically, $\tilde{y}$ is from $k$ classes most confusable with the ground-truth label of $x_0$, as predicted by a classifier $f$ trained on the source dataset.
By default, we take the highest-scoring member as follows.
\begin{equation}
\label{eq:cfg_whole}
\begin{aligned}
\tilde{y} \;=&\; \arg\max_{c \in \mathcal{C}_k(x_0)} f(x_0),\\ \mathcal{C}_k(x) \;=& \;\operatorname{TopK}\!\big(f(x_0),\,k\big),
\end{aligned}
\end{equation}
where $f(x_0)$ denotes the predicted logit for class $c$, $\mathcal{C}_k(x_0)$ is the index set of the top-$k$ predicted classes for $x_0$ and $k=\big\lfloor C/w\big\rfloor$. $C$ is the total class number and $w$ is the stable weight.
By setting an appropriate $w$, the generator can supply abundant near-boundary targets for rebalancing extremely imbalanced data while avoiding large semantic shifts that would distort the feature distribution.

%本文提出的基于diffusion的对抗样本生成由加噪去原类特定信息和去噪加目标类特定信息两个过程组成，如图5所示。在加噪扩散过程中，为了去除类原始的类特定信息，我们首先模拟以当前类为目标的一步去噪过程，由模型预测从当前xt到去噪后的干净x0的噪声估计，如下：
%公式
%在得到噪声估计后按前向闭式将xt加噪到更大的时间步，如下：
%公式
%如此，便可以模仿对抗性攻击的去类特定信息过程。特别的，这种加噪过程共执行T/10步，以避免生成样本与原始类特征中心偏移过远，达到生成分类边界附近的对抗性样本的目的。此外，受到常用对抗性攻击更关注于深层特征的启发，为了改变样本的深层类特定信息，以达到稳定攻击的目的（减少攻击失败和生成类特定信息偏移的现象），加噪时间步t从T/2开始。

%在获得所需的去除类特定信息的噪声图像xnoised后，在去噪过程中，xnoised被重新加上标准cfg引导，引导到其近似类，在达到生成类边界附近样本目标的同时，避免生成样本跨度态度，严重影响特征分布空间，公式如下：
%公式
%其中受deepfool的启发【】，近似类的标签由源数据集训练的分类器输出的logits得出，表示如下：
%公式
%这里为了尽可能为极端不平衡数据集提供更多边界样本的同时减少生成样本的类特定信息偏移程度导致的分类空间被进一步扰乱的现象，top被设为numclasses/3。
% \usepackage{multirow}
\begin{table*}[]
\caption{Top-1 accuracy (\%) of ResNet-32 on CIFAR100-LT across baselines. CBDM* represents CBDM-based. Improvements on tail classes and the overall average are highlighted in red. AVG computes the overall accuracy and improvement of each baseline.}
\label{tab:agr-cifar100}
\centering
\setlength{\tabcolsep}{4pt}
% \rowcolors{0}{gray!20}
\rowcolors{3}{gray!10}{white}
\vskip -8pt
\begin{tabular}{c|cccc|cccc|cccc|c}
\toprule
\rule{0pt}{2.3ex}
\multirow{2}{*}{\textbf{Method}} &
  \multicolumn{4}{c|}{\textbf{Imbalanced ratio-10}} &
  \multicolumn{4}{c|}{\textbf{Imbalanced ratio-100}} &
  \multicolumn{4}{c|}{\textbf{Imbalanced ratio-200}}&\multirow{2}{*}{\textbf{Avg}} \\
 &
  \multicolumn{1}{c}{\textbf{Head}} &
  \multicolumn{1}{c}{\textbf{Med}} &
  \multicolumn{1}{c}{\textbf{Tail}} &
  \multicolumn{1}{c|}{\textbf{All}} &
  \multicolumn{1}{c}{\textbf{Head}} &
  \multicolumn{1}{c}{\textbf{Med}} &
  \multicolumn{1}{c}{\textbf{Tail}} &
  \multicolumn{1}{c|}{\textbf{All}} &
  \multicolumn{1}{c}{\textbf{Head}} &
  \multicolumn{1}{c}{\textbf{Med}} &
  \multicolumn{1}{c}{\textbf{Tail}} &
  \multicolumn{1}{c|}{\textbf{All}} \\ \toprule
\rule{0pt}{2.3ex}
\textbf{CBDM*} &
  \multicolumn{1}{c}{68.17} &
  \multicolumn{1}{c}{61.80} &
  55.40 &
  \multicolumn{1}{c|}{61.79} &
  66.30 &
  52.85 &
  25.93 &
  48.81 &
  69.10 &
  46.52 &
  15.60 &
  44.02 &51.54\\
\textbf{CBDM*+DBG} &
  67.77 &
  63.67 &
  \textcolor{myred}{\textbf{57.20}} &
  62.96 &
  70.08 &
  53.35 &
  \textcolor{myred}{\textbf{26.72}} &
  50.37 &
  69.17 &
  49.15 &
  \textcolor{myred}{\textbf{18.67}} &
  46.01& \textcolor{myred}{↑1.57}\\
\textbf{CBDM} &
  \multicolumn{1}{c}{66.87} &
  \multicolumn{1}{c}{64.28} &
  59.17 &
  \multicolumn{1}{c|}{63.52} &69.10
   & 53.75
   & 28.60
   & 50.81
   & 64.10
   & 48.98
   & 20.67
   & 45.02&53.12
   \\
\textbf{CBDM+DBG} & 68.37
   & 63.62
   & \textcolor{myred}{\textbf{59.27}}
   & 63.74
   & 67.30
   & 54.98
   & \textcolor{myred}{\textbf{30.10}}
   & 51.21
   & 64.40
   & 49.27
   & 20.47
   & 45.17&\textcolor{myred}{↑0.25}
   \\
\textbf{OCLT} &
  \multicolumn{1}{c}{68.60} &
  \multicolumn{1}{c}{62.92} &
  56.80 &
  62.79 &
  68.70 &
  52.38 &
  25.73 &
  49.28 &
  69.03 &
  48.90 &
  16.23 &
  45.14&52.40 \\
\textbf{OCLT+DBG} &
  67.93 &
  63.58 &
  \textcolor{myred}{\textbf{57.10}} &
  62.94 &
  68.20 &
  53.95 &
  \textcolor{myred}{\textbf{27.37}} &
  50.25 &
  69.33 &
  49.33 &
  \textcolor{myred}{\textbf{16.33}} &
  45.43&\textcolor{myred}{↑0.47} \\
\textbf{{DiffuLT}} &70.27
   &63.30
   &54.30
   &62.69
   &68.10
   &51.20
   &29.37
   &49.72
   &68.67
   &49.02
   &19.13
   &45.95&52.79
   \\
\textbf{{DiffuLT+DBG}} &68.33
   &63.00
   &\textcolor{myred}{\textbf{56.53}}
   &62.66
   &68.87
   &54.55
   &28.63
   &51.07
   &69.30
   &49.52
   &\textcolor{myred}{\textbf{19.43}}
   &46.43&\textcolor{myred}{↑0.60}
   \\ \toprule
\end{tabular}
\vskip -10pt
\end{table*}

\begin{table*}[]
\caption{Top-1 accuracy (\%) of ResNet-32 on CIFAR10-LT across baselines. CBDM* represents CBDM-based.}
\label{tab:agr-cifar10}
\centering
\setlength{\tabcolsep}{4pt}
\rowcolors{3}{gray!10}{white}
\vskip -8pt
\begin{tabular}{c|cccc|cccc|cccc|c}
\toprule
\rule{0pt}{2.3ex}
\multirow{2}{*}{\textbf{Method}} &
  \multicolumn{4}{c|}{\textbf{Imbalanced ratio-10}} &
  \multicolumn{4}{c|}{\textbf{Imbalanced ratio-100}} &
  \multicolumn{4}{c|}{\textbf{Imbalanced ratio-200}}&
  \multirow{2}{*}{\textbf{AVG}}\\
 &
  \multicolumn{1}{c}{\textbf{Head}} &
  \multicolumn{1}{c}{\textbf{Med}} &
  \multicolumn{1}{c}{\textbf{Tail}} &
  \multicolumn{1}{c|}{\textbf{All}} &
  \multicolumn{1}{c}{\textbf{Head}} &
  \multicolumn{1}{c}{\textbf{Med}} &
  \multicolumn{1}{c}{\textbf{Tail}} &
  \multicolumn{1}{c|}{\textbf{All}} &
  \multicolumn{1}{c}{\textbf{Head}} &
  \multicolumn{1}{c}{\textbf{Med}} &
  \multicolumn{1}{c}{\textbf{Tail}} &
  \multicolumn{1}{c|}{\textbf{All}}
  \\ \toprule
  \rule{0pt}{2.3ex}
\textbf{CBDM*} &
  \multicolumn{1}{c}{94.60} &
  \multicolumn{1}{c}{85.92} &
  \multicolumn{1}{c}{88.17} &
  \multicolumn{1}{c|}{89.20} &
  91.90 &
  76.20 &
  76.53 &
  81.01 &
  92.10 &
  72.10 &
  65.30 &
  76.06 &82.09\\
\textbf{CBDM*+DBG} &
  94.20 &
  86.42 &
  88.13 &
  89.27 &
  92.13 &
  77.62 &
  74.97 &
  81.18 &
  91.30 &
  74.17 &
  \textcolor{myred}{\textbf{69.00}} &
  77.76& \textcolor{myred}{↑0.65}\\
\textbf{CBDM} &
  \multicolumn{1}{c}{93.77} &
  \multicolumn{1}{c}{86.22} &
  \multicolumn{1}{c}{90.87} &
  \multicolumn{1}{c|}{89.88} &92.73
   & 77.08
   & 76.30
   & 81.54
   & 91.90
   & 73.38
   & 67.27
   & 77.10&82.84
   \\
\textbf{CBDM+DBG}   & 93.00
   &86.17
   &90.33
   &89.47
   &93.60
   &77.45
   &\textcolor{myred}{\textbf{76.53}}
   &82.02
   &90.73
   &71.95
   &\textcolor{myred}{\textbf{68.77}}
   &76.63&\textcolor{mygreen}{↓0.13}
   \\
\textbf{OCLT} &
  \multicolumn{1}{c}{93.93} &
  \multicolumn{1}{c}{86.47} &
  \multicolumn{1}{c}{90.10} &
  89.80 &
  92.50 &
  77.65 &
  77.43 &
  82.04 &
  92.10 &
  74.50 &
  68.70 &
  78.04&83.29 \\
\textbf{OCLT+DBG} &
  93.23 &
  86.10 &
  \textcolor{myred}{\textbf{90.57}} &
  89.58 &
  91.60 &
  78.90 &
  \textcolor{myred}{\textbf{78.90}} &
  82.71 &
  91.70 &
  75.45 &
  \textcolor{myred}{\textbf{70.87}} &
  78.95&\textcolor{myred}{↑0.46} \\
\textbf{DiffuLT} &93.07
   &85.53
   &89.70
   &89.04
   &92.47
   &77.15
   &76.33
   &81.50
   &93.93
   &72.40
   &62.23
   &77.56&82.7
   \\
\textbf{DiffuLT+DBG} &93.07
   &86.47
   &89.43
   &89.34
   &92.20
   &76.85
   &\textcolor{myred}{\textbf{76.73}}
   &81.42
   &92.23
   &73.97
   &\textcolor{myred}{\textbf{69.93}}
   &78.24&\textcolor{myred}{↑0.30}
   \\ \toprule
\end{tabular}
\vskip -10pt
\end{table*}

\begin{table*}[]
\caption{Top-1 accuracy (\%) of ViT-B/16 on CIFAR100-LT across baselines. CBDM* represents CBDM-based.}
\label{tab:vitagr-cifar100}
\centering
\setlength{\tabcolsep}{4pt}
\rowcolors{3}{gray!10}{white}
\vskip -8pt
\begin{tabular}{c|cccc|cccc|cccc|c}
\toprule
\rule{0pt}{2.3ex}
\multirow{2}{*}{\textbf{Method}} &
  \multicolumn{4}{c|}{\textbf{Imbalanced ratio-10}} &
  \multicolumn{4}{c|}{\textbf{Imbalanced ratio-100}} &
  \multicolumn{4}{c|}{\textbf{Imbalanced ratio-200}}&\multirow{2}{*}{\textbf{Avg}} \\
 &
  \multicolumn{1}{c}{\textbf{Head}} &
  \multicolumn{1}{c}{\textbf{Med}} &
  \multicolumn{1}{c}{\textbf{Tail}} &
  \multicolumn{1}{c|}{\textbf{All}} &
  \multicolumn{1}{c}{\textbf{Head}} &
  \multicolumn{1}{c}{\textbf{Med}} &
  \multicolumn{1}{c}{\textbf{Tail}} &
  \multicolumn{1}{c|}{\textbf{All}} &
  \multicolumn{1}{c}{\textbf{Head}} &
  \multicolumn{1}{c}{\textbf{Med}} &
  \multicolumn{1}{c}{\textbf{Tail}} &
  \multicolumn{1}{c|}{\textbf{All}} \\ \toprule
\rule{0pt}{2.3ex}
\textbf{CBDM*} &
  \multicolumn{1}{c}{49.73} &
  \multicolumn{1}{c}{42.45} &
  \multicolumn{1}{c}{32.73} &
  \multicolumn{1}{c|}{41.72} &
  45.17 &
  27.77 &
  9.83 &
  27.61 &
  43.80 &
  23.93 &
  5.73 &
  24.43&31.25 \\
\textbf{CBDM*+DBG} &49.60
   &43.75
   &\textcolor{myred}{\textbf{36.50}}
   &
  43.33 &47.97
   &30.02
   &\textcolor{myred}{\textbf{11.07}}
   &
  29.72 &
   46.10&
   26.88&
   \textcolor{myred}{\textbf{6.37}}&
  26.49&\textcolor{myred}{↑1.93} \\
\textbf{CBDM} &
  \multicolumn{1}{c}{50.57} &
  \multicolumn{1}{c}{47.38} &
  \multicolumn{1}{c}{40.07} &
  \multicolumn{1}{c|}{46.14} &48.17
   & 34.20
   & 16.30
   & 33.02
   & 45.97
   & 29.95
   & 10.20
   & 28.83&36.00
   \\
\textbf{CBDM+DBG} & 51.60
   & 47.48
   & 39.37
   & 46.28
   & 49.77
   & 34.85
   & 15.63
   & 33.56
   & 46.20
   & 29.55
   & 9.83
   & 28.63&\textcolor{myred}{↑0.16}
   \\
\textbf{OCLT} &
  \multicolumn{1}{c}{50.93} &
  \multicolumn{1}{c}{45.67} &
  \multicolumn{1}{c}{36.17} &
  44.40 &
  47.73 &
  31.18 &
  11.70 &
  30.30 &45.77
   &25.23
   &6.03
   &25.63&33.44
   \\
\textbf{OCLT+DBG} &
  50.23 &
  45.75 &
  \textcolor{myred}{\textbf{37.63}} &
  44.66 &
  49.13 &
  32.05 &
  \textcolor{myred}{\textbf{12.37}} &
  31.27 &
  47.97 &
  27.27 &
  \textcolor{myred}{\textbf{6.70}} &
  27.31&\textcolor{myred}{↑0.97} \\
\textbf{{DiffuLT}} &50.93
   &44.20
   &34.03
   &43.17
   &49.43
   &33.30
   &14.93
   &32.63
   &47.57
   &27.30
   &5.70
   &26.90&34.23
   \\
\textbf{{DiffuLT+DBG}} &50.90
   &45.52
   &\textcolor{myred}{\textbf{38.57}}
   &45.05
   &49.23
   &34.58
   &\textcolor{myred}{\textbf{15.43}}
   &33.23
   &50.17
   &30.73
   &\textcolor{myred}{\textbf{10.27}}
   &30.42&\textcolor{myred}{↑2.00}
   \\ \toprule
\end{tabular}
\vskip -10pt
\end{table*}

\subsection{Classifier-Driven Bifurcated Data-cleaning}

Due to generator bias that inherit from long-tailed dataset and the feature space mismatch between generator and classifier, the generation process may produce misleading adversarial samples with bad class-specific features. Such hard samples cannot provide precise boundary supervision for the classifier and may even further damage the decision boundaries. To address this, we propose a long-tailed-insensitive classifier-driven bifurcated data-cleaning pipeline, as shown in~\autoref{fig:filter}. The pipeline comprises two parallel data-cleaning branches: one employs prototype-distance filtering to discard outlier samples and boundary-distant samples, while the other applies confidence-credibility filtering to remove samples misaligned with the feature space of the classifier.

For prototype-distance filtering, a well-trained classifier $f_\theta(\cdot)$ is first obtained by training on the source long-tailed dataset \(\mathcal{D}\) with logit adjustment loss~\cite{menonlong}. Then, for each class \(c\in\{1,\dots,C\}\), we compute its prototype as follows.
\begin{equation}
\begin{aligned}
\mu_k = \frac{1}{N_k}\sum_{(x_i,\,y_i=c)\in\mathcal{D}} \tilde{z}(x_i),\quad
\mu_k = \frac{\mu_k}{\|\mu_k\|_2},
\label{eq:proto_center}
\end{aligned}
\end{equation}
where $\tilde{z}(x) = \frac{\phi(x)}{\|\phi(x)\|_2}$, \(\phi(\cdot)\) is feature extractor of \(f_\theta\) and $\tilde{z}(x)$ denotes the normalized feature. After that we can measure the cosine distance $d_c(\cdot)$ of each generated sample $\hat{x_0}$ to their target class prototype as follows.
\begin{align}
d_c\big(\hat{x_0}\big) \;&=\; 1 - \big\langle \tilde{z}(\hat{x_0}),\, \mu_k \big\rangle,
\label{eq:cosine_dist}
\end{align}
where $\big\langle \tilde{z}(\hat{x_0}),\, \mu_k \big\rangle$ represents cosine similarity between vector $\tilde{z}(\hat{x_0})$ and $\mu_k$. Then, to better fit the long-tailed distributions, we obtain an highest distance $d_c^h$ and lowest distance $d_c^l$ and define an class-wise acceptance range as $[(1-l)d_c^l, (1+h)d_c^h]$ for each class. Both $h$ and $l$ ratio parameters that are set to small values to remove extreme outlier samples and boundary-distant samples.

For confidence-credibility filtering, the confidence and credibility of the classifier for $\hat{x_0}$ are computed following~\autoref{eq:confidence} with disturbing class set as the second highest predicted class using $f_\theta(\cdot)$. Then, to mitigate the influence of inherited long-tailed bias in the classifier on data filtering, the class-wise threshold $a$ is defined according to the total samples of each class. A sample is removed only when it is misclassified into non-original and non-target labels with high confidence and credibility. By using long-tail-insensitive classifier-driven bifurcated data-cleaning pipeline to filter samples, generated samples that harmful to the classifier's decision boundaries can be effectively removed. 
After data-cleaning, the accepted adversarial set $S_{adv}$ can work as a plug-and-play auxiliary training set to refine the feature space boundary of the classifier as 
$
\mathcal{D}_{\mathrm{aug}} 
\;=\; 
\mathcal{D} \;\cup\; \big\{\,\big(x,\,\tilde{y}\big)\;:\; x \in S_{adv} \,\big\}.
\label{eq:augment_set}
$ By providing additional boundary information for long-tailed datasets, we alleviate the boundary ambiguity induced by class-imbalance bias and prevent further blurring caused by feature leakage in head-to-tail transfer strategies.

%由于对抗性攻击的方向和目标类没有被施加严格细粒度的方向控制，且diffusion特征空间与分类器特征空间存在差异，生成过程可能存在攻击失败和攻击偏移的现象。这些噪声样本不仅无法为特征提取backbone提供细粒度的分类边界指导，还可能进一步扰乱特征空间。为了解决该问题，我们进一步提出了双重数据筛选策略。如图6所示，双重数据筛选策略由类原型距离筛选和类置信度筛选机制组成。
%类原型距离筛选机制主要是为了过滤无用样本和偏离样本，以避免由于diffusion特征空间与分类器特征空间差异造成的干扰问题。具体的，使用原数据集和LA策略训练的分类器首先从原数据集中提取每个类的原型中心，表示如下：
%公式
%在提取到原型中心后，计算生成样本与类原型距离，筛掉过近的无用样本和过远的偏离样本，如下：【这块需要看代码进一步加长】
%公式
%类置信度筛选机制主要是为了过滤由于特征空间复杂导致的攻击偏移的样本，即类特定信息偏移到未知类的样本。具体的，分类器对所有生成样本输出logits，当分类置信度较高时，我们认为分类器能正确分类该样本。如果分类结果与对抗性攻击前的类标签和攻击目标标签都不同，则认为该样本发生攻击偏移。整体流程表示如下：
%公式
%经过双重数据筛选后，生成的样本被加入训练集，用于补充缺失样本的同时，优化特征分布空间，减轻类间分布混淆。
\section{Experiments}
\label{sec:Experiments}

\begin{table*}[]
\caption{Top-1 accuracy (\%) of ViT-B/16 on CIFAR10-LT across baselines. CBDM* represents CBDM-based.}
\label{tab:vitagr-cifar10}
\centering
\setlength{\tabcolsep}{4pt}
% \rowcolors{0}{gray!20}
\rowcolors{3}{gray!10}{white}
\vskip -8pt
\begin{tabular}{c|cccc|cccc|cccc|c}
\toprule
\rule{0pt}{2.3ex}
\multirow{2}{*}{\textbf{Method}} &
  \multicolumn{4}{c|}{\textbf{Imbalanced ratio-10}} &
  \multicolumn{4}{c|}{\textbf{Imbalanced ratio-100}} &
  \multicolumn{4}{c|}{\textbf{Imbalanced ratio-200}}&
  \multirow{2}{*}{\textbf{AVG}}\\
 &
  \multicolumn{1}{c}{\textbf{Head}} &
  \multicolumn{1}{c}{\textbf{Med}} &
  \multicolumn{1}{c}{\textbf{Tail}} &
  \multicolumn{1}{c|}{\textbf{All}} &
  \multicolumn{1}{c}{\textbf{Head}} &
  \multicolumn{1}{c}{\textbf{Med}} &
  \multicolumn{1}{c}{\textbf{Tail}} &
  \multicolumn{1}{c|}{\textbf{All}} &
  \multicolumn{1}{c}{\textbf{Head}} &
  \multicolumn{1}{c}{\textbf{Med}} &
  \multicolumn{1}{c}{\textbf{Tail}} &
  \multicolumn{1}{c|}{\textbf{All}}
  \\ \toprule
  \rule{0pt}{2.3ex}
\textbf{CBDM*} &
  \multicolumn{1}{c}{77.50} &
  \multicolumn{1}{c}{68.38} &
  \multicolumn{1}{c}{78.17} &
  \multicolumn{1}{c|}{74.05} &
  65.07 &
  54.05 &
  66.63 &
  61.13 &
  67.77 &
  47.55 &
  52.93 &
  55.23 &63.47\\
\textbf{CBDM*+DBG} &72.67
   &70.50
   &\textcolor{myred}{\textbf{80.83}}
   &
  74.25 &
  60.57 &
  56.88 &
  \textcolor{myred}{\textbf{72.17}} &
  63.32 &
  66.13 &
  51.38 &
  \textcolor{myred}{\textbf{57.60}} &
  57.67&\textcolor{myred}{↑1.61} \\
\textbf{CBDM} &
  \multicolumn{1}{c}{66.97} &
  \multicolumn{1}{c}{72.75} &
  \multicolumn{1}{c}{83.83} &
  \multicolumn{1}{c|}{74.34} &64.80
   & 61.83
   & 72.60
   & 65.95
   & 57.73
   & 52.08
   & 68.47
   & 58.69&66.33
   \\
\textbf{CBDM+DBG}   & 66.40
   &73.53
   &\textcolor{myred}{\textbf{84.73}}
   &74.75
   &65.63
   &61.05
   &\textcolor{myred}{\textbf{74.43}}
   &66.44
   &59.77
   &49.42
   &\textcolor{myred}{\textbf{70.47}}
   &58.84&\textcolor{myred}{↑0.35}
   \\
\textbf{OCLT} &
  \multicolumn{1}{c}{68.40} &
  \multicolumn{1}{c}{71.83} &
  \multicolumn{1}{c}{84.30} &
  74.54 &
  61.80 &
  59.40 &
  57.40 &
  62.14 &
  58.60 &
  52.10 &
  60.00 &
  56.42&64.37 \\
\textbf{OCLT+DBG} &
  70.03 &
  70.40 &
  \textcolor{myred}{\textbf{85.27}} &
  74.75 &
  60.40 &
  57.62 &
  \textcolor{myred}{\textbf{73.17}} &
  63.12 &
  61.97 &
  52.85 &
  \textcolor{myred}{\textbf{64.43}} &
  59.06&\textcolor{myred}{↑1.27} \\
\textbf{DiffuLT} &
  \multicolumn{1}{c}{74.07} &
  \multicolumn{1}{c}{68.20} &
  \multicolumn{1}{c}{77.83} &
  72.85 &
  61.80 &
  54.92 &
  70.30 &
  61.60 &
  73.67 &
  55.08 &
  59.17 &
  61.88&65.44
   \\
\textbf{DiffuLT+DBG} &
  71.13 &
  70.70 &
  \textcolor{myred}{\textbf{83.50}} &
  74.67 &
  59.87 &
  57.73 &
  \textcolor{myred}{\textbf{73.00}} &
  62.95 &
  68.70 &
  59.95 &
  \textcolor{myred}{\textbf{62.13}} &
  63.23&\textcolor{myred}{↑1.51}
   \\ \toprule
\end{tabular}
\vskip -10pt
\end{table*}

% \begin{table*}[]
% \caption{cifar.}
% \label{tab:sin-cifar10}
% \centering
% \begin{tabular}{c|ccc|ccc|ccc}
% \toprule
% \rule{0pt}{2.3ex}
% \multirow{2}{*}{Method} & \multicolumn{3}{c|}{CIFAR10-LT} & \multicolumn{3}{c|}{CIFAR100-LT} & \multicolumn{3}{c}{Statisics} \\
%            & 10   & 100  & 200 & 10   & 100  & 200 & Head & Med  & Tail \\ \toprule
%            \rule{0pt}{2.3ex}
% CUDA       & -    & -    & -   & 58.4 & 47.6 &   -  & 67.3 & 50.4 & 21.4 \\
% CSA        & \textbf{90.8} & 82.5 &   -  & 62.6 & 46.6 &   -  & 64.3 & 49.7 & 18.2 \\
% ADRW       & 90.3 & \textbf{83.6} &     & 61.9 & 46.4 &     & -    & -    & -    \\
% H2T        & -    & -    &   -  & -    & 48.9 &  -   & -    & -    & -    \\
% CBDM-based &  89.2    &   81.0   &   76.0  &   61.7   &   48.8   &   44.0  &   66.3   &   52.8   &   25.9   \\
% CBDM       &   \underline{89.8}   &   81.5   &   77.1  &   \textbf{63.5}   &   \textbf{50.8}   &   45.0  &   69.1   &   \textbf{53.7}   &   28.6   \\
% OCLT  & \underline{89.8}    &   \underline{82.0}   &   \textbf{78.0}   &   62.7  &   49.2   &   45.1   &   68.7  &   52.3   &   25.7      \\
% DiffuLT    &   89.0   &   81.5   &   77.5  &   62.6   &   49.7   &   \textbf{45.9}  &   68.1   &   51.2   &   \textbf{29.3}   \\
% DBG  &   88.6   &   81.8   &   77.2  &   62.8   &   50.4   &   45.1   &   \textbf{70.1}  &   53.4   &    26.7  \\ 
% \toprule
% \end{tabular}
% \end{table*}

\subsection{Experimental Setup}
\label{sec:setup}
\noindent\textbf{Datasets.} We evaluate DBG along with other baselines on two long-tailed datasets, CIFAR10-LT and CIFAR100-LT~\cite{cui2019class}, constructed following \citet{cao2019learning}. The imbalance ratio is set to 10, 100, and 200 to assess the proposed DBG under varying levels of class imbalance.

\noindent\textbf{Baselines.} Comparisons are made to diffusion-based generative methods for long-tailed learning, including CBDM-based~\cite{qin2023class}, CBDM (with an additional loss), OCLT~\cite{zhang2024long}, and DiffuLT~\cite{shao2024diffult}. 
Experiments for CBDM-based, CBDM, and OCLT use the authors' released implementations. 
DiffuLT is re-implemented from the paper due to the absence of official code.

\noindent\textbf{Implementation.} We set $w=3$, $l=0.02$, and $h=0.05$ for CIFAR datasets. $a$ is linearly decayed from 0.9 to 0.5. Specially, to better estimate the drift of the feature space, we sweep $\lambda$ over five values uniformly spaced in $[2.5, 3.0]$ and report the mean. 
For all methods, we use ResNet-32~\cite{he2016deep} as the classifier backbone and train for 200 epochs with a batch size of 128. The initial learning rate is $0.1$ and is multiplied by $0.1$ at epochs 160 and 180. 
To better validate the universality of DBG, we also report results with ViT-B/16~\cite{DBLP:conf/iclr/DosovitskiyB0WZ21}, trained for 100 epochs with a batch size of 32, a base learning rate of $1\times10^{-4}$, and a weight decay of $0.01$.

\begin{table}[]
\caption{Ablation study on DBG using ResNet-32. CBDM* represents CBDM-based (without our framework) and the first line is standard learning without any augmentation.}
\label{tab:ablation}
\centering
\setlength{\tabcolsep}{4pt}
\vskip -8pt
\begin{tabular}{ccc|cccl}
\toprule
\rule{0pt}{2.3ex}
\textbf{Gen.} & \textbf{PD.} &\textbf{ CC.} & \textbf{Head}  & \textbf{Med}   & \textbf{Tail}  & \textbf{All}   \\ \toprule
\rule{0pt}{2.3ex}
     &     &     & 64.1 & 35.6 & 8.6 & 36.1 \\
CBDM* & & &66.3&52.8&25.9&48.8\\
\checkmark    &     &     & 66.4 & 50.5 & 23.2 & 47.0 \\
\checkmark    & \checkmark   &     & \underline{69.2} & \underline{52.7} & \textbf{27.6} & \underline{50.1} \\
\checkmark    &    & \checkmark   & 68.4 & 48.9 & 23.5 & 47.1 \\
\checkmark    & \checkmark   & \checkmark   & \textbf{70.1} & \textbf{53.4} & \underline{26.7} & \textbf{50.4} \\ \toprule
\end{tabular}
\vskip -10pt
\end{table}

\subsection{Classification Results}
\label{sec:results}

To test whether DBG-generated samples reduce decision-boundary ambiguity and improve long-tailed recognition, auxiliary-generation retraining is conducted on CIFAR-LT with two classifier backbones (ResNet-32 and ViT-B/16).
In each baseline, the data generated by DBG is used as a plug-and-play auxiliary training set in addition to the generated data of baseline. 
For ResNet-32, per-group accuracies on head, medium, tail splits and the overall score are reported in~\autoref{tab:agr-cifar100} and~\ref{tab:agr-cifar10}. 
Across methods, adding DBG consistently increases overall accuracy, with larger gains on tail classes.
These results indicate that near-boundary samples from DBG consistently enhance class separability, mitigate head-class bias, and thus improve long-tailed learning. 

Similar experiments are conducted for ViT-B/16 as shown in~\autoref{tab:vitagr-cifar100} and~\ref{tab:vitagr-cifar10}), where all baselines also improve after injecting data generated by DBG, indicating that the complementary boundary information supplied by DBG generalizes across architectures. It is noted that tail class accuracy drops slightly on CIFAR100-LT with CBDM. This is because CBDM uses a strong tail-bias loss that makes tail classes easier to classify. Even so, DBG improves the overall accuracy, indicating a better decision space. Moreover, consistent gains across baselines imply that boundary ambiguity is pervasive across data augmentation methods, with head-to-tail transfer further amplifying the problem.

%为了更好的验证所提方法生成的数据集对恢复特征空间分布边界和提供额外信息的能力，我们分别进行了辅助生成重训练实验和独立生成重训练实验。
%辅助生成重训练。我们将DBG生成的数据集作为其他方法的辅助集用于重训练分类器，具体的，通过在其他方法生成的数据集中插入百分之30的DBG生成的数据集，用于提供额外边界信息重建更可分且尾部类偏移更小的平衡分类空间。在表1和表2中分别展示了基于CIFAR10-LT和CIFAR100-LT数据集的结合不同方法辅助生成重训练后头，中，尾部类和总的分类精度变化。可以看出在插入DBG生成的数据集后，所有方法的总体精度都有提升，尤其是尾部类提升最为明显，这说明DBG生成的数据集提供的额外边界分布知识确实帮助分类器学习到一个更可分的特征空间，同时减轻了分类器对头部类的偏向，有效提高了长尾学习的效果。
%独立生成重训练实验。DBG生成的数据集单独用于重平衡长尾训练集其他方法对比，用于检测它提供缺失尾部类知识的能力。在表3和表4中分别展示了基于CIFAR10-LT和CIFAR100-LT数据集的重训练分类器后头，中，尾部类和总的分类精度变化。值得注意的是，虽然大部分时候DBG并没取得最优的分类精度，但是这是因为DBG注重于生成位于分类边界的对抗性样本，这使得当独立使用其重平衡数据集时对比起其他方法会出现接近类原型的额外特征的缺乏，然而在大多数条件下它都取得了不错的效果，证实了其提供额外特征信息的能力。

\begin{table}[]
\caption{Hyperparameter sensitivity study on $h$ and $l$.}
\label{tab:hl}
\centering
\setlength{\tabcolsep}{5pt}
\vskip -8pt
\begin{tabular}{c|ccccccc}
\toprule
\rule{0pt}{2.3ex}
\textbf{$h$}       & 0.00 & 0.05          & 0.05          & 0.05 & 0.05          & 0.07 & 0.10 \\
\textbf{$l$}       & 0.02 & 0.00          & 0.02          & 0.05 & 0.10          & 0.02 & 0.02 \\ \toprule \rule{0pt}{2.3ex}
\textbf{Head}    & 69.4 & \textbf{70.1} & \textbf{70.1} & 68.8 & 69.3          & 68.5 & 68.8 \\
\textbf{Med}     & 52.7 & 52.5          & \textbf{53.4} & 52.1 & 52.2          & 52.2 & 52.0 \\
\textbf{Tail}    & 25.7 & 27.6          & 26.7          & 27.5 & \textbf{27.7} & 27.5 & 26.2 \\
\textbf{Overall} & 49.6 & 50.3          & \textbf{50.4} & 49.7 & 50.0          & 49.6 & 49.3 \\ \hline
\end{tabular}
\vskip -10pt
\end{table}

\subsection{Ablation Study}
\label{sec:ablation}
On CIFAR100-LT with imbalance ratio set to 100, the effect of removing the prototype-distance and confidence-credibility filters is evaluated via ablations in which the classifier is retrained from scratch. As shown in \autoref{tab:ablation}, dropping either filter causes a substantial decrease in accuracy, underscoring the importance of these filtering mechanisms. Because the diffusion model in DBG is trained following the CBDM-based setting, results are also reported against CBDM-based. Moreover, because accuracy drops sharply when the prototype-distance filter is removed, we study the sensitivity of $h$ and $l$. As shown in~\autoref{tab:hl}, small value changes of $h$ and $l$ do not materially affect DBG’s performance in long-tailed learning, and we already set them to their optimal values.

\begin{figure}
\centering
{\includegraphics[width=0.45\textwidth]{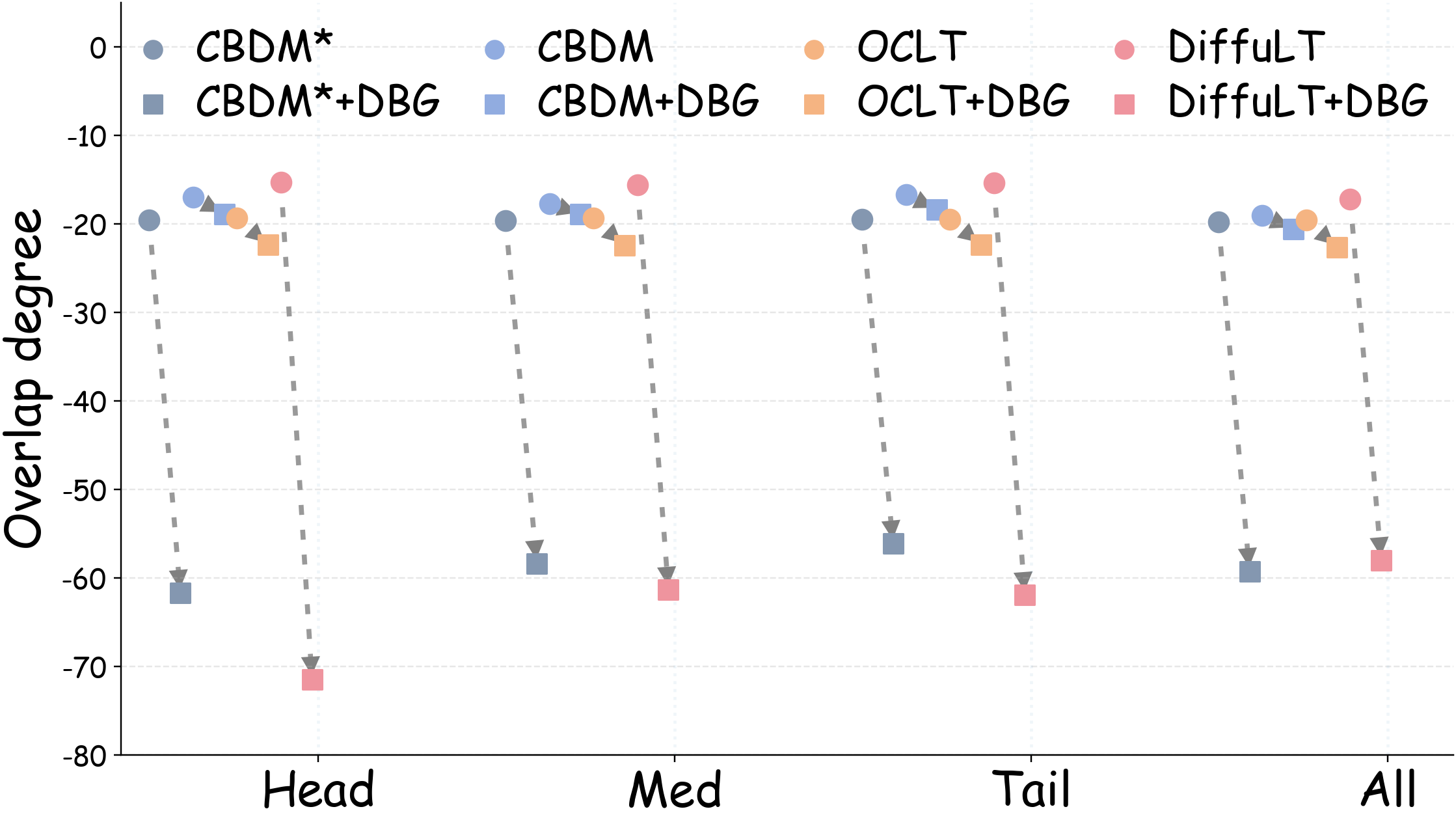}}
\vspace{-10pt}
\caption{Inter-class overlap changes across baselines with DBG on ResNet-32 with imbalanced rate set to 100 for CIFAR100-LT.}
\vskip -8pt
\label{fig:quan}
\end{figure}

\begin{figure}
\centering
{\includegraphics[width=0.45\textwidth]{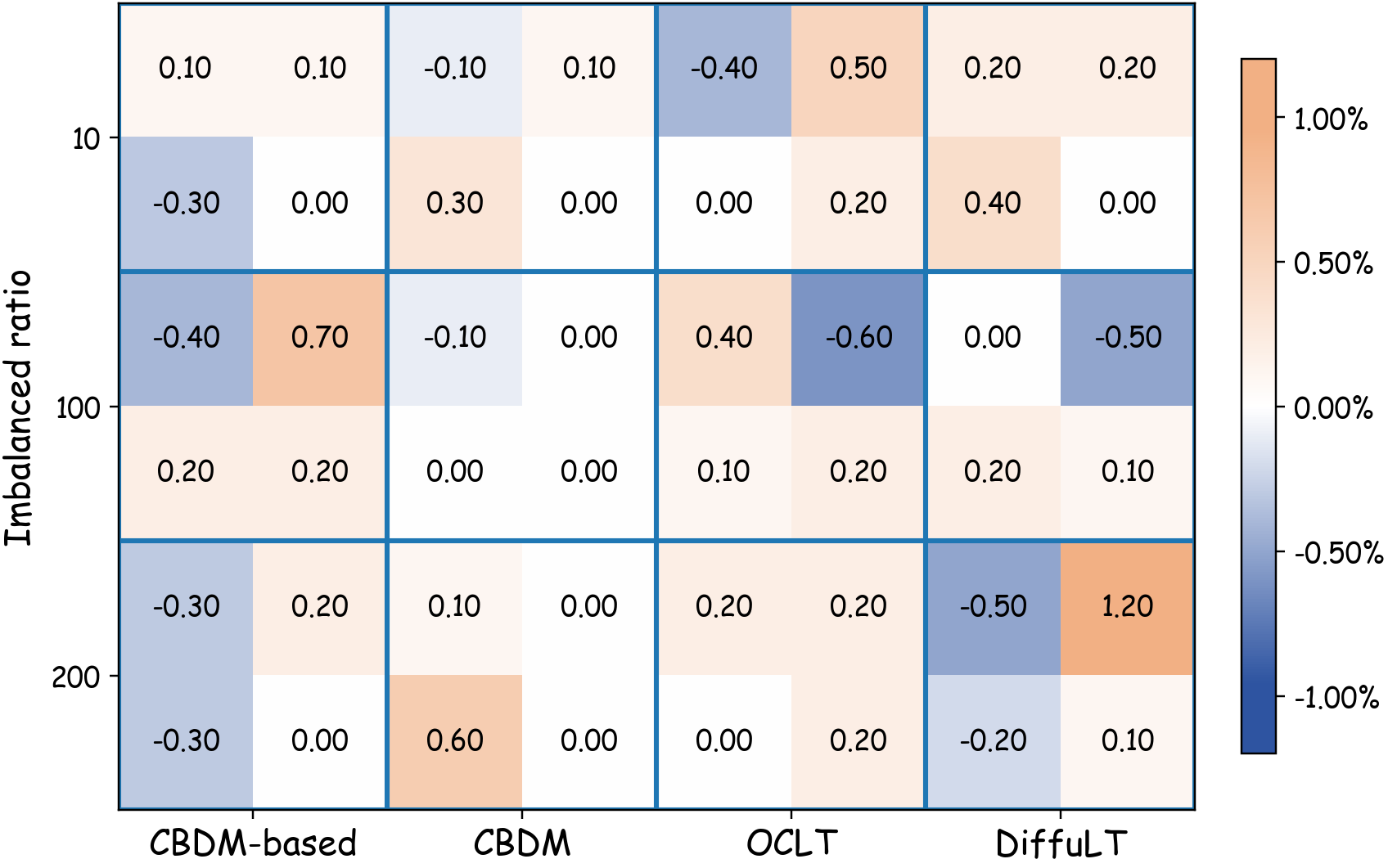}}
\vspace{-10pt}
\caption{Outlier-rate changes across baselines with DBG. In each $2\times 2$ tile, quadrants correspond to head (top-left), medium (top-right), tail (bottom-left), and overall (bottom-right).}
\vskip -13pt
\label{fig:quanr}
\end{figure}

% 需要在导言区启用：
% \usepackage{subcaption}

%在对比学习中，我们在CIFAR-LT数据集上测试了去除原型距离筛选和置信度筛选后重训练分类器的分类精度变化。如表5所示，在去除任意一种筛选机制后，重训练分类器的分类精度都发生了大幅度的下降，证明了筛选机制的有效性。
%此外，我们还分别对参数w，h，l和a/n_c进行参数敏感度实验，结果如表6所示。可以看出过高或过低的参数设置会较大影响生成数据集对长尾学习的有效性，但在合理范围内这些参数的改变是相对不敏感的。进一步说明了DBG的有效性。

\subsection{Quantitative Analysis}
\label{sec:ablation}
To directly evaluate the ability of our method in reconstructing clear decision boundaries and balanced feature distributions, we augment the training set with DBG-generated samples and measure decision boundary quality using the proposed overlap degree and the outlier rate.
As shown in~\autoref{fig:quan}, incorporating DBG-generated data reduces overall inter-class overlap across baselines, indicating that DBG helps form a more separable decision space. In most cases, the overlap degree of tail class is also reduced, suggesting that the method mitigates long-tailed bias by providing additional information to underrepresented classes. Similar effects are observed when the imbalance ratio is set to 10 and 200 shown in the appendix.
As shown in~\autoref{fig:quanr}, DBG reduces the overall outlier rate across baselines, suggesting that an improved decision space helps balance the the classifier bias. In a few cases, outlier rates for some classes increase slightly, likely because DBG focuses on boundary refinement, causing minor perturbations to specific classes.

\begin{figure}
\centering
{\includegraphics[width=0.47\textwidth]{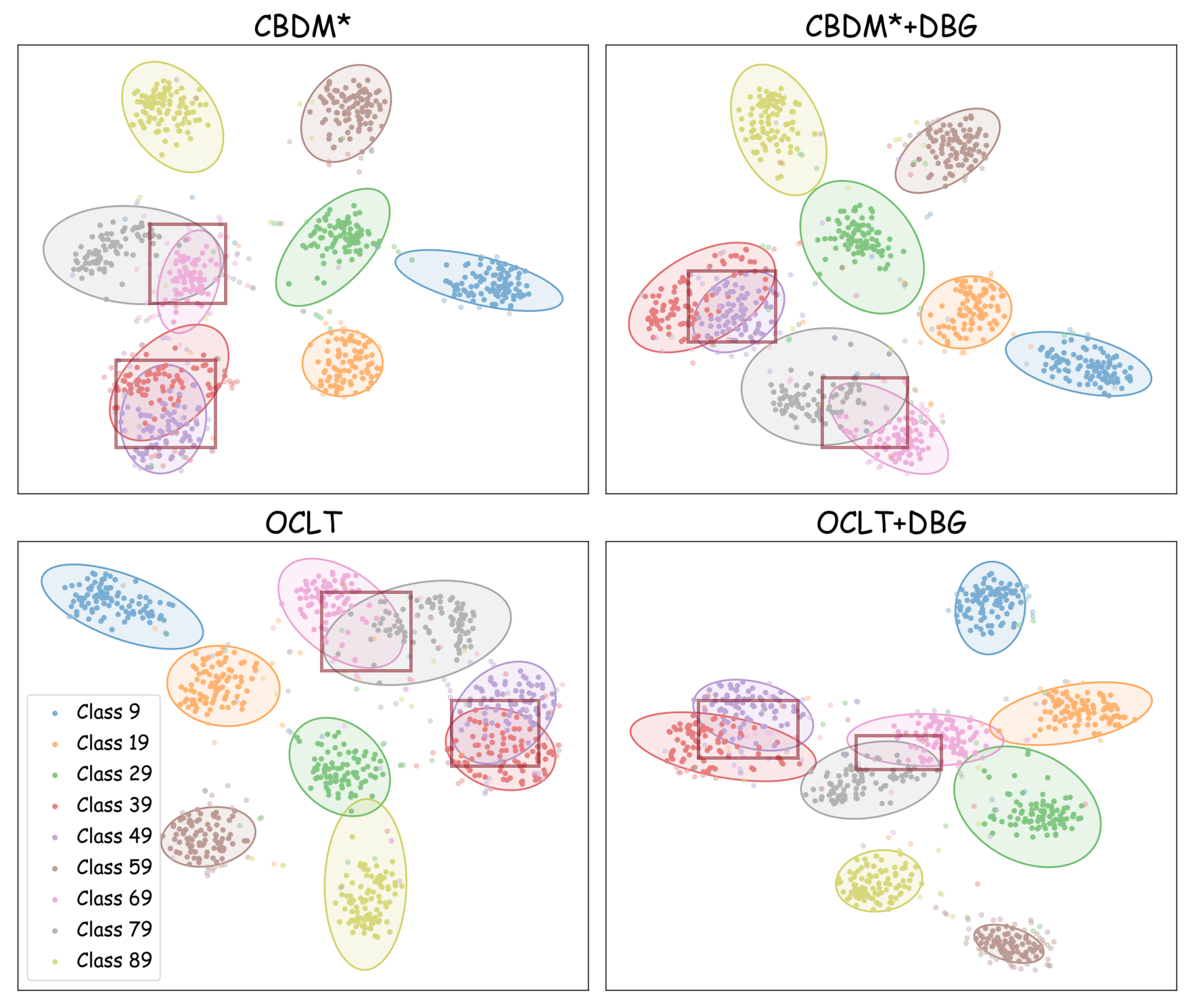}}
\vspace{-8pt}
\caption{Changes in Feature space under different baselines.}
\vskip -13pt
\label{fig:tsne}
\end{figure}

Moreover, to more intuitively demonstrate the sharpening effect of DBG on the classification boundaries, t-SNE visualizations based on different baselines are plotted, as shown in~\autoref{fig:tsne}. It can be observed that after injecting DBG-generated data, the inter-class overlap in the decision space of different baselines decreases, while the inter-class distances increase and the intra-class features become more compact, which verifies the effectiveness of GBT in addressing the boundary ambiguity problem.
%为了直接评估我们的方法重建清晰决策边界和修复特征分布的能力，我们用DBG生成的样本增强训练集，并报告了通过我们所提出的重叠程度指标和离群点率两个指标测试的特征分布边界的变化。如图7所示，在使用了DBG生成的数据集后，各个方法的总类间重叠度都有所下降证实了我们方法恢复可分决策空间的能力。且在大部分情况下我们的方法都使得尾部类分类空间与其他类分布重叠更小了，这也说明我们所提出的方法对长尾偏向是有帮助的，能潜在的提供更多缺失的尾部类信息。
%如图8所示，在使用了DBG生成的数据集后，各个方法的总离群点率都有所下降，尤其是尾部类，说明我们的方法通过构建更好决策空间的做法确实能够拉回偏移的尾部类特征分布。另外值得注意的是，在一些情况下，头部类和中部类的离群点率出现增高的现象，这是因为我们的方法并没有提供太多的头部类边界样本来辅助训练，同时在增强尾部类边界可分性的同时，头部类难免受到挤压，但从整体上看，这种轻微的头部类偏移带来了巨大的尾部类的恢复，所以是可以接受的。

\section{Conclusion}
\label{sec:conclusion}
Long-tailed recognition is hindered not only by data scarcity but also by boundary ambiguity introduced by data augmentation and head-to-tail transfer in generative augmentation. We made boundary ambiguity measurable with three metrics: inter-class overlap degree, outlier rate, and generation confidence. Building on these findings, we proposed Decision Boundary-aware Generation (DBG), which generates near-boundary samples and removes harmful ones through a classifier-driven bifurcated cleaning pipeline to offer additional decision boundary information for long-tailed learning. Across long-tailed benchmarks, DBG reduces inter-class overlap and outliers, and improves tail and overall accuracy, yielding a more uniform and more separable decision space. However, long-tailed bias in the diffusion still reduced the effectiveness of DGB-generated samples.
Going forward, we will study adaptive cleaning and fine-tuning based method adaptation to further restore long-tailed decision boundaries.

\section*{Acknowledgment}
This study was supported in part by the National Natural Science Foundation of China under Grants 62376233, 62502402 and 62306181; in part by the Natural Science Foundation of Fujian Province under Grant 2024J09001; in part by the Guangdong Basic and Applied Basic Research Foundation under Grant 2024A1515010163; in part by the Shenzhen Science and Technology Program under Grant RCBS20231211090659101; in part by the National Key Laboratory of Radar Signal Processing  under Grant JKW202403; in part by Xiaomi Young Talents Program and YGZ was funded by Inno HK Generative AI R\&D Center.
{
    \small
    \bibliographystyle{ieeenat_fullname}
    \bibliography{cite}
}
% \input{LaTeXAuthor_Guidelines_for_Proceedings/sec/X_suppl}


\end{document}